# Joint Training Across Multiple Activation Sparsity Regimes

Haotian Wang
Wanght816@zju.edu.cn

**Affiliation**
Department of Nuclear Medicine and PET-CT Center, the Second Affiliated Hospital, Zhejiang University School of Medicine, Hangzhou, Zhejiang, China

**Abstract**

Generalization in deep neural networks remains only partially understood. Inspired by the stronger generalization tendency of biological systems, we explore the hypothesis that robust internal representations should remain effective across both dense and sparse activation regimes. To test this idea, we introduce a simple training strategy that applies global top-k constraints to hidden activations and repeatedly cycles a single model through multiple activation budgets via progressive compression and periodic reset. Using CIFAR-10 without data augmentation and a WRN-28-4 backbone, we find in single-run experiments that two adaptive keep-ratio control strategies both outperform dense baseline training. These preliminary results suggest that joint training across multiple activation sparsity regimes may provide a simple and effective route to improved generalization.

## 1. Introduction

Generalization is one of the core goals of modern neural network research. In overparameterized models trained mainly by empirical risk minimization, there are often many parameter configurations that can fit the training set. Existing evidence shows that deep networks can fit not only real labels but also random labels or noisy targets, indicating that model capacity or conventional empirical risk minimization alone is insufficient to explain practical generalization[1]. At the same time, both explicit and implicit regularization, the optimization path, and training dynamics may jointly shape the final generalization performance[2].

In contrast to mainstream artificial neural networks, biological nervous systems often exhibit stronger generalization and a greater tendency to resist overfitting under limited data and complex environments. From a biologically inspired perspective, revisiting how highly generalizable internal solutions emerge may provide useful guidance for network design[3], [4]. Based on this motivation, we propose an exploratory hypothesis: during early learning, a system may depend more on relatively dense activation patterns to establish task-relevant representations; as learning becomes more mature, the system may further migrate toward a more energy-efficient and compact sparse activation state. In other words, representations that remain stable under both dense and sparse activation conditions may be closer to solutions with better generalization.

Sparsity has been extensively studied, but most prior work has focused on weight pruning, dropout, sparse autoencoders, or specialized routing mechanisms. In comparison, relatively fewer studies trained a model across multiple activation rates within supervised learning. Classical work has shown that k-sparse mechanisms can effectively construct sparse representations[5]. Meanwhile, rectifier networks such as ReLU naturally exhibit a high proportion of zero activations, suggesting that activation sparsity itself is an important modeling dimension worthy of separate investigation[6].

Motivated by these considerations, we adopt a deliberately simple method that remains compatible with standard training pipelines. Specifically, we introduce a global top-k activation constraint during convolutional network training and repeatedly expose the same model to different activation budgets through progressive compression and periodic reset. We then examine whether this form of joint training across multiple activation sparsity regimes can improve generalization.

## 2. Method

### 2.1 Dataset and Task Setup

We conduct image classification experiments on the CIFAR-10 dataset using the standard official train/test split[7]. To minimize the influence of additional explicit regularization on result interpretation, no data augmentation is used. In particular, we do not apply random cropping, random flipping, or similar transformations. Only tensor conversion and standard normalization are performed. Input images are normalized channel-wise using mean values of (0.4914, 0.4822, 0.4465) and standard deviations of (0.2470, 0.2435, 0.2616). The batch size is set to 128.

To improve reproducibility, the random seed is fixed at 3407, and the random states of Python, NumPy, and PyTorch are all initialized accordingly.

### 2.2 Backbone Architecture

The backbone model is a Wide Residual Network (WRN) [8], specifically WRN-28-4, with depth 28 and widen factor 4. No dropout is used. This architecture is chosen for two main reasons. First, WRN allows convenient control of model capacity by adjusting network width. Second, residual connections help preserve information flow when local activations are suppressed, which improves training stability after introducing sparse activation constraints.

The network consists of an initial 3 x 3 convolution, followed by three wide residual groups, each containing four basic blocks, and a final classification head. Each basic block contains two convolutional layers and adopts a pre-activation design in which normalization and activation sparsity control are applied before convolution. After the final residual group, one additional normalization layer and one activation sparsity control layer are applied, followed by global average pooling and a fully connected layer that outputs logits for 10 classes.

### 2.3 Normalization Strategy

We use RMSNorm2d as the primary normalization method instead of BatchNorm[9]. For a convolutional feature map x in $R^{N \times C \times H \times W}$, normalization is performed at each spatial location along the channel dimension. Concretely, the root-mean-square value is computed as:
$RMS(x) = \sqrt{mean(x^2) + epsilon}$

The input is then scaled at each location without mean-centering. In implementation, the normalization layer includes a learnable scale parameter and a learnable bias term, with epsilon set to 1e-10. This design is intended to reduce the extra regularization introduced by batch statistics, making it easier to isolate the contribution of activation sparsity control to generalization.

### 2.4 Activation Sparsity Control via Top-k

We introduce a hard top-k sparsity constraint at multiple activation sites in the network. Each sparse activation module first applies a nonlinear activation and then performs top-k selection. In the current implementation, the activation function is ReLU. Negative values are therefore first clipped to zero, after which sparsity selection is applied to the remaining positive activations.

For a feature map x in $R^{N \times C \times H \times W}$, we use topk_mode = global. For each sample, the activation tensor is flattened across the (C, H, W) dimensions, yielding $D = C \times H \times W$ total elements. Given the current keep ratio r, only the largest ceil(r x D) activations are retained, while all other locations are set to zero. Because ReLU has already removed negative responses, the top-k operator effectively preserves only the strongest positive activations in that layer.

This top-k constraint is applied at multiple locations throughout the network, including both activation sites within each residual block and one activation site before the final classifier head. Therefore, the sparsity constraint is not limited to a single layer, but instead acts on multiple intermediate representations across the backbone.

**2.5 Optimization and Training Setup**

The model is trained with SGD using an initial learning rate of 0.1 and Nesterov momentum. The main optimization settings are as follows: momentum = 0.9, weight decay = 0, batch size = 128, training epochs = 500.

The learning rate is scheduled with cosine annealing over the full 500 epochs. The training objective is standard cross-entropy loss.

**2.6 Adaptive Keep-Ratio Controllers**

To alternately train the same model under different activation budgets, we evaluate two adaptive keep-ratio controllers that dynamically adjust the keep ratio r of all top-k modules on an epoch-by-epoch basis.

Strategy 1 (additive compression with local drop-triggered reset).
Training starts from r = 1. At each epoch, r is reduced by a fixed additive step of 0.01. Training accuracy is smoothed using an exponential moving average with factor 0.9. When the smoothed training accuracy decreases by 0.01 relative to the previous epoch, the model is considered to have entered an overly sparse regime. At that point, r is reset to 1, and the same process is repeated.

Strategy 2 (multiplicative compression with best-gap-triggered reset).
Training also starts from r = 1. At each epoch, r is multiplied by 0.98, corresponding to a fixed multiplicative decay. Training accuracy is smoothed using an exponential moving average with factor 0.5. When the smoothed training accuracy falls more than 0.2 below the best historical training accuracy, the current sparsity level is considered too strong. The keep ratio is then reset to 1, and the cycle repeats.

In practice, these controllers create a periodic compression-recovery-recompression loop. During training, the model is repeatedly driven from relatively dense activation states to more sparse states and then returned to dense states again, forcing it to learn representations that remain functional across multiple activation budgets.

**3. Results**

Without additional activation compression, the dense baseline model achieves a best test accuracy of 0.869 (Fig.1).

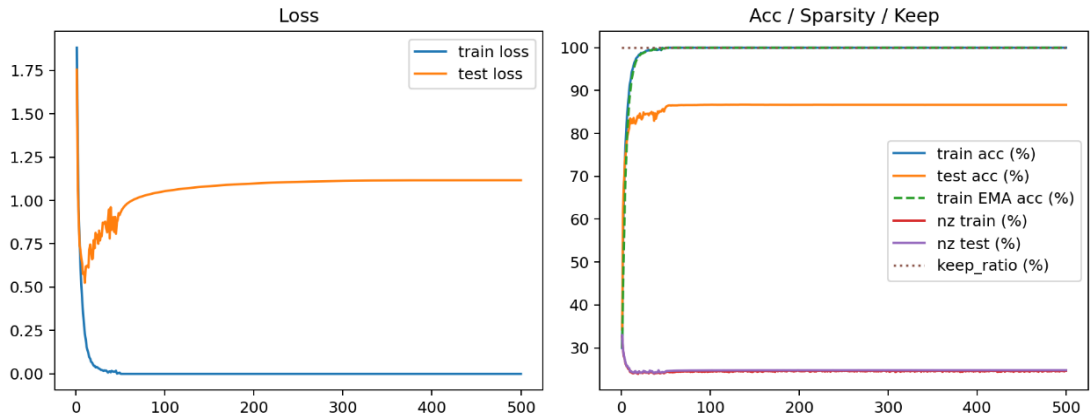

**Fig.1** Baseline mode. acc = accuracy, nz = non-zero.

After introducing top-k-based activation sparsity with Strategy 1, the best test accuracy improves to 0.8797, with the peak performance observed at epoch 295 (Fig.2).

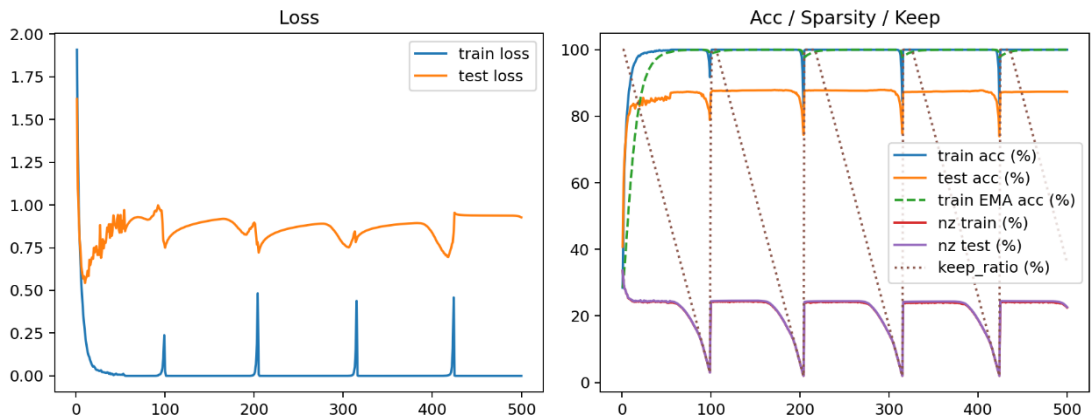

**Fig.2** Strategy 1 training. acc = accuracy, nz = non-zero.

With Strategy 2, the best test accuracy further reaches 0.8802, with the best result observed at epoch 164 (Fig.3).

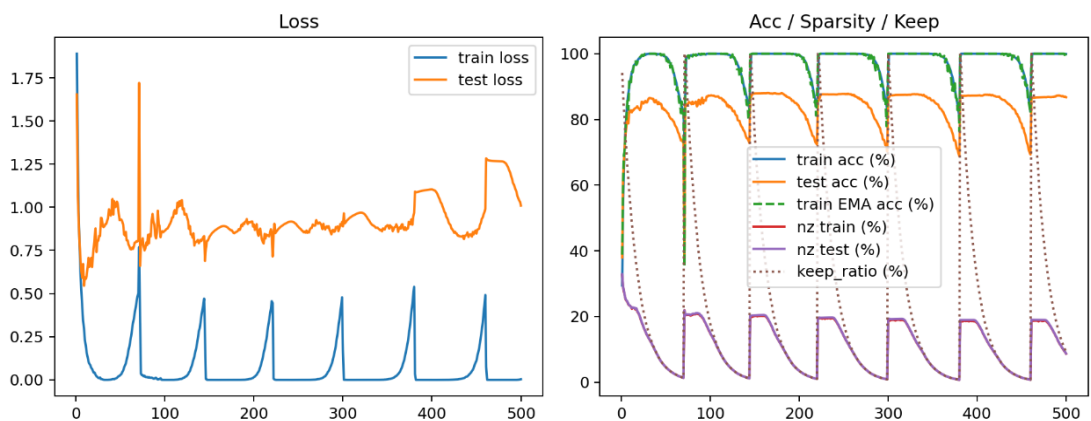

**Fig.3** Strategy 2 training. acc = accuracy, nz = non-zero.

In single-run experiments, both adaptive activation sparsity strategies outperform the dense baseline.

## 4. Discussion

This study adopts a simple and highly training-compatible strategy that explicitly controls neuronal activation rates through a top-k rule and progressively compresses the activation budget during training, thereby forcing the same model to operate repeatedly under different activation regimes. Preliminary results suggest that this approach may improve test performance, indicating that "placing pressure on activations and requiring the model to adapt to multiple sparsity levels" may be a promising direction for enhancing generalization.

Two observations are broadly consistent with our working hypothesis. First, the training scheme appears to show some resistance to overfitting. Even without data augmentation, models trained with activation compression outperform the dense baseline on the test set. Second, when the model has sufficient capacity, performance does not collapse even when the effective activation rate is compressed to relatively low levels. This suggests that the network may contain a substantial compressible activation space, meaning that some activations are not always necessary for final discrimination.

At the same time, we observe several non-trivial phenomena that merit further study. Most notably, the best generalization does not emerge during continuous compression, but rather after the activation budget has been restored to a relatively high level. This suggests that simply making the network sparser does not directly translate into better generalization. Instead, the alternation between sparse-constrained phases and dense-recovery phases may jointly encourage the model to converge to a more robust parameter solution.

In addition, even when the extra top-k constraint is relaxed, the network still naturally maintains a relatively low true activation rate under ReLU. This is consistent with prior observations that rectifier networks exhibit inherently sparse activations[6]. It also implies that future analyses should distinguish between the nominal keep ratio and the actual nonzero activation rate, since failing to do so may overestimate the effective strength of the external sparsity constraint.

From a methodological perspective, activation sparsity may be more dynamic and reversible than weight sparsity. Weight pruning directly alters the parameter-space structure, whereas activation sparsity primarily constrains information flow during forward propagation. As a result, it is easier to switch between sparse and dense regimes within the same trained model. This property may make activation sparsity especially useful for studying how structural constraints during training affect generalization, rather than serving only as a deployment-time compression tool.

## 5. Limitations

This work is only a preliminary biologically inspired exploration and has several limitations.

First, hyperparameters have not yet been systematically optimized. The present results should therefore be interpreted as proof of concept rather than as a performance upper bound.

Second, from a biological perspective, the compression process may not ideally be driven by standard backpropagation. A more biologically plausible formulation would likely involve a feedforward-like adaptation mechanism. At present, we have not identified an effective feedforward alternative.

Third, due to hardware limitations, the current experiments have not been extended to larger-scale models or to tasks beyond image classification. It remains possible that this framework may show different or even stronger effects in other settings, such as reinforcement learning or large language models.

## 6. Conclusion

We present a simple top-k-based activation sparsity training framework that repeatedly exposes a single model to multiple activation budgets through adaptive compression and periodic reset. In single-run CIFAR-10 experiments without data augmentation, both tested controllers improve upon dense baseline training. These preliminary findings suggest that joint training across multiple activation sparsity regimes may help promote better generalization. Further work is needed to validate this effect across repeated runs, broader architectures, larger-scale settings, and more biologically plausible adaptation mechanisms.